\documentclass[10pt,twocolumn,letterpaper]{article}

\usepackage{iccv}
\usepackage{times}
\usepackage{epsfig}
\usepackage{graphicx}
\usepackage{amsmath}
\usepackage{amssymb}
\usepackage{booktabs}
\usepackage{multirow}
\usepackage{subcaption}
\usepackage{array}
\newcolumntype{M}[1]{>{\centering\arraybackslash}m{#1}}

\newcommand{\residual}[2]{
  $\left[
    \begin{array}{cc}
      3 \times 3 \times 3, #1 \\
      3 \times 3 \times 3, #1
    \end{array}
  \right]
  \times #2$
}

\usepackage[pagebackref=true,breaklinks=true,letterpaper=true,colorlinks,bookmarks=false]{hyperref}

\iccvfinalcopy 

\def\iccvPaperID{19} 

\ificcvfinal\fi
\begin{document}

\title{Learning Spatio-Temporal Features with 3D Residual Networks \\ for Action Recognition}

\author{Kensho Hara, Hirokatsu Kataoka, Yutaka Satoh\\
National Institute of Advanced Industrial Science and Technology (AIST)\\
Tsukuba, Ibaraki, Japan\\
{\ttfamily\small \{kensho.hara, hirokatsu.kataoka, yu.satou\}@aist.go.jp}
}

\maketitle

\begin{abstract}
  Convolutional neural networks with spatio-temporal 3D kernels (3D CNNs)
  have an ability to directly extract spatio-temporal features from videos for action recognition.
  Although the 3D kernels tend to overfit because of a large number of their parameters,
  the 3D CNNs are greatly improved by using recent huge video databases.
  However, the architecture of 3D CNNs is relatively shallow against to
  the success of very deep neural networks in 2D-based CNNs, such as residual networks (ResNets).
  In this paper, we propose a 3D CNNs based on ResNets toward a better action representation.
  We describe the training procedure of our 3D ResNets in details.
  We experimentally evaluate the 3D ResNets on the ActivityNet and Kinetics datasets.
  The 3D ResNets trained on the Kinetics did not suffer from overfitting
  despite the large number of parameters of the model,
  and achieved better performance than relatively shallow networks, such as C3D.
  Our code and pretrained models (e.g. Kinetics and ActivityNet) are publicly available at
  \href{https://github.com/kenshohara/3D-ResNets}{https://github.com/kenshohara/3D-ResNets}.
\end{abstract}

\section{Introduction}
  One important type of real-world information is human actions.
  Automatically recognizing and detecting human action in videos are widely used in applications
  such as surveillance systems, video indexing, and human computer interaction.

  Convolutional neural networks (CNNs) achieve high performance in action recognition \cite{I3D,Feichtenhofer16,Simonyan2014,C3D}.
  Most of the CNNs use 2D convolutional kernels \cite{STResNet,Feichtenhofer16,Simonyan2014,Wang2015TDD,Wang2016TSN}, similar to the CNNs for image recognition.
  The two-stream architecture \cite{Simonyan2014} that consists of RGB and optical flow streams
  is often used to represent spatio-temporal information in videos.
  Combining the both streams improves action recognition performance.

  Another approach that captures the spatio-temporal information adopts spatio-temporal 3D convolutional kernels \cite{I3D,Ji2013,C3D} instead of the 2D ones.
  Because of the large number of parameters of the 3D CNNs,
  training them on relatively small video datasets, such as UCF101 \cite{UCF101} and HMDB51 \cite{HMDB51},
  leads to lower performance compared with the 2D CNNs pretrained on large-scale image datasets, such as ImageNet \cite{imagenet_cvpr09}.
  Recent large-scale video datasets, such as Kinetics \cite{Kinetics}, greatly contribute to improve the recognition performance of the 3D CNNs \cite{I3D,Kinetics}.
  The 3D CNNs are competitive to the 2D CNNs even though
  their architectures are relatively shallow compared with the architectures of 2D CNNs .

  Very deep 3D CNNs for action recognition have not been explored enough
  because of the training difficulty caused by the large number of their parameters.
  Prior work in image recognition shows very deep architectures of CNNs improves recognition accuracy \cite{ResNet,Inception}.
  Exploring various deeper models for the 3D CNNs and achieving lower loss at convergence are important to improve action recognition performance.
  Residual networks (ResNets) \cite{ResNet} are one of the most powerful architecture.
  Applying the architecture of ResNets to 3D CNNs is expected to contribute further improvements of action recognition performance.

  In this paper, we experimentally evaluate 3D ResNets to get good models for action recognition.
  In other words, the goal is to generate a standard pretrained model in spatio-temporal recognition.
  We simply extend from the 2D-based ResNets to the 3D ones.
  We train the networks using the ActivityNet and Kinetics datasets and evaluate their recognition performance.

  Our main contribution is exploring the effectiveness of ResNets with 3D convolutional kernels.
  We expect that this work gives further advances to action recognition using 3D CNNs.

\section{Related Work}
  We here introduce action recognition databases and approaches.

  \subsection{Action Recognition Database}
    The HMDB51 \cite{HMDB51} and UCF101 \cite{UCF101} are the most successful databases in action recognition.
    The recent consensus, however, tells that these two databases are not large-scale databases.
    It is difficult to train good models without overfitting using these databases.
    More recently, huge databases such as Sports-1M \cite{KarpathyCVPR14} and YouTube-8M \cite{YouTube8M} are proposed.
    These databases are big enough whereas their annotations are
    noisy and only video-level labels (i.e. the frames that do not relate to target activities are included).
    Such noise and unrelated frames might prevent models from good training.
    In order to create a successful pretrained model like 2D CNNs trained on ImageNet \cite{imagenet_cvpr09},
    the Google DeepMind released the Kinetics human action video dataset \cite{Kinetics}.
    The Kinetics dataset includes 300,000 or over trimmed videos and 400 categories.
    The size of Kinetics is smaller than Sports-1M and YouTube-8M
    whereas the quality of annotation is extremely high.

    We use the Kinetics in order to optimize 3D ResNets.

  \subsection{Action Recognition Approach}
    One of the popular approach of CNN-based action recognition is two-stream CNNs with 2D convolutional kernels.
    Simonyan et al. proposed the method that uses RGB and stacked optical flow frames as appearance and motion information, respectively \cite{Simonyan2014}.
    They showed combining the two-streams improves action recognition accuracy.
    Many methods based on the two-stream CNNs are proposed to improve action recognition performance \cite{STResNet,Feichtenhofer16,Wang2015TDD,Wang2016TSN}.
    Feichtenhofer et al. proposed combining two-stream CNNs with ResNets \cite{Feichtenhofer16}.
    They showed the architecture of ResNets is effective for action recognition with 2D CNNs.
    Different from the above mentioned approaches, we focused on 3D CNNs, which recently outperform the 2D CNNs using large-scale video datasets.

    Another approach adopts CNNs with 3D convolutional kernels.
    Ji et al. proposed to apply the 3D convolution to extract spatio-tepmoral features from videos.
    Tran et al. trained 3D CNNs, called C3D, using the Sports-1M dataset \cite{KarpathyCVPR14}.
    They experimentally found $3 \times 3 \times 3$ convolutional kernel achieved best performance.
    Varol et al. showed expanding temporal length of inputs for 3D CNNs improves recognition performance \cite{LongTermTemporalConv}.
    They also found using optical flows as inputs to 3D CNNs outperforms RGB inputs and combining RGB and optical flows achieved best performance.
    Kay et al. showed the results of 3D CNNs on their Kinetics dataset are competitive to the results of 2D CNNs pretrained on ImageNet
    whereas the results of 3D CNNs on the UCF101 and HMDB51 are inferior to the results of the 2D CNNs.
    Carreira et al. introduced the inception architecture \cite{Inception}, which is very deep network (22 layers), to the 3D CNNs
    and achieved state-of-the-art performance \cite{I3D}.
    In this paper, we introduce the ResNet architecture, which outperforms the inception architecture in image recognition, to the 3D CNNs.

\section{3D Residual Networks}
  \subsection{Network Architecture}
\begin{figure}[t]
      \centering
      \includegraphics[width=0.95\linewidth, clip]{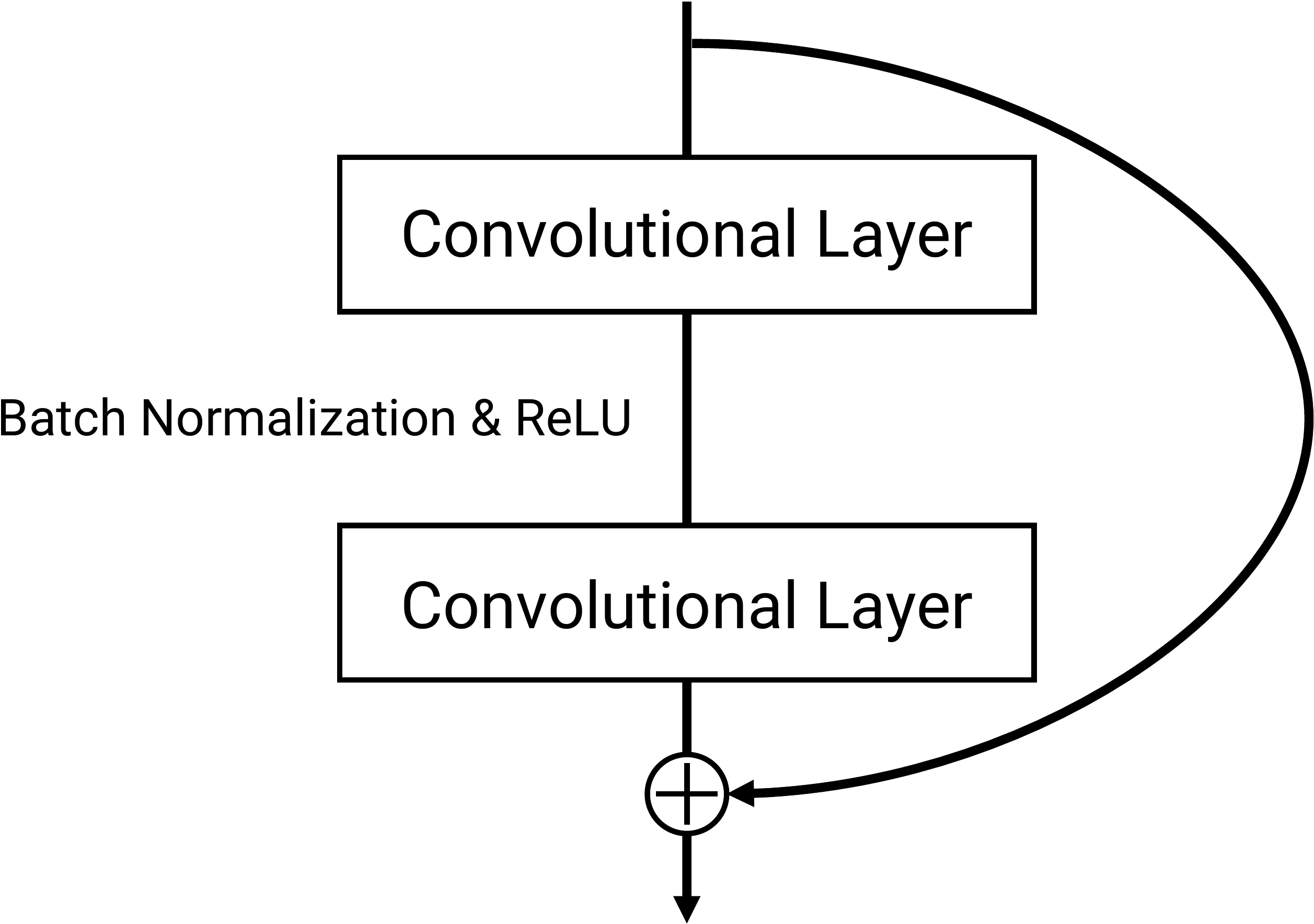}
      \caption{
        Residual block.
        Shortcut connections bypass a signal from the top of the block to the tail.
        Signals are summed at the tail.
      }
      \label{fig:residual}
    \end{figure}

    Our network is based on ResNets \cite{ResNet}.
    ResNets introduce shortcut connections that bypass a signal from one layer to the next.
    The connections pass through the gradient flows of networks from later layers to early layers,
    and ease the training of very deep networks.
    Figure \ref{fig:residual} shows the residual block, which is an element of ResNets.
    The connections bypass a signal from the top of the block to the tail.
    ResNets are conssits of multiple residual blocks.

    Table \ref{tbl:network} shows our network architecture.
    The difference between our networks and original ResNets \cite{ResNet} is the number of dimensions of convolutional kernels and pooling.
    Our 3D ResNets perform 3D convolution and 3D pooling.
    The sizes of convolutional kernels are $3 \times 3 \times 3$,
    and the temporal stride of conv1 is 1, similar to C3D \cite{C3D}.
    The network uses 16 frame RGB clips as inputs.
    The sizes of input clips is $3 \times 16 \times 112 \times 112$.
    Down-sampling of the inputs is performed by conv3\_1, conv4\_1, conv5\_1 with a stride of 2
    When the number of feature maps increased, we adopt identity shortcuts with zero-padding (type A in \cite{ResNet}) to avoid increasing the number of parameters.

    \begin{table*}[t]
        \centering
        \captionsetup{width=0.65\linewidth}
        \caption{
          Network Architecture.
          Residual blocks are shown in brackets.
          Each convolutional layer is followed by batch normalization \cite{BatchNorm} and ReLU \cite{ReLU}.
          Down-sampling is performed by conv3\_1, conv4\_1, conv5\_1 with a stride of 2.
          The dimension of last fully-connected layer is set for the Kinetics dataset (400 categories).
        }
        \label{tbl:network}
        \begin{tabular}{ccc}
            \toprule
            \multirow{2}{*}{Layer Name} & \multicolumn{2}{c}{Architecture} \\
            \cmidrule{2-3}
            & 18-layer & 34-layer \\ \midrule
            conv1 & \multicolumn{2}{c}{$7 \times 7 \times 7, 64$, stride 1 (T), 2 (XY)} \\ \midrule
            \multirow{2}{*}{conv2\_x} & \multicolumn{2}{c}{$3 \times 3 \times 3$ max pool, stride 2} \\
            & \residual{64}{2} & \residual{64}{3} \\ \midrule
            conv3\_x & \residual{128}{2} & \residual{128}{4} \\ \midrule
            conv4\_x & \residual{256}{2} & \residual{256}{6} \\ \midrule
            conv5\_x & \residual{512}{2} & \residual{512}{3} \\ \midrule
            & \multicolumn{2}{c}{average pool, 400-d fc, softmax} \\
            \bottomrule
        \end{tabular}
    \end{table*}

  \subsection{Implementation}
    \subsubsection{Training}
      We use stochastic gradient descent (SGD) with momentum to train our network.
      We randomly generate training samples from videos in training data to perform data augmentation.
      We first select temporal positions of each sample by uniform sampling.
      16 frame clips are generated around the selected temporal positions.
      If the videos are shorter than 16 frames, we loop the videos as many times as necessary.
      We then randomly selects the spatial positions from the 4 corner or 1 center, similar to \cite{VeryDeepTwo}.
      In addition to the positions, we also select the spatial scales of each sample to perform multi-scale cropping \cite{VeryDeepTwo}.
      The scales are selected from $\left\{1, \frac{1}{2^{1/4}}, \frac{1}{\sqrt{2}}, \frac{1}{2^{1/4}}, \frac{1}{2}\right\}$.
      The scale $1$ means a maximum scale (i.e. the size is the length of short side of frame).
      The aspect ratio of cropped frame is $1$.
      The generated samples are horizontally flipped with 50\% probability.
      We also perform mean subtraction for each sample.
      All generated samples have the same class labels as their original videos.

      To train the 3D ResNets on the Kinetics dataset, we use SGD with a mini-batch size of 256 on 4 GPUs (NVIDIA TITAN X) using the training samples described above.
      The weight decay is 0.001 and the momentum is 0.9.
      We start from learning rate 0.1,
      and divide it by 10 for three times after the validation loss saturates.
      In preliminary experiments on the ActivityNet dataset, large learning rate and batch size was important to achieve good recognition performance.

    \subsubsection{Recognition}
      We recognize actions in videos using the trained model.
      We adopt the sliding window manner to generate input clips, (i.e. each video is split into non-overlapped 16 frame clips.)
      Each clip is cropped around a center position with the maximum scale.
      We estimate class probabilities of each clip using the trained model,
      and average them over all clips of a video to recognize actions in videos.

\section{Experiments}
  \begin{figure*}[t]
    \begin{subfigure}[b]{0.5\linewidth}
      \includegraphics[width=\linewidth]{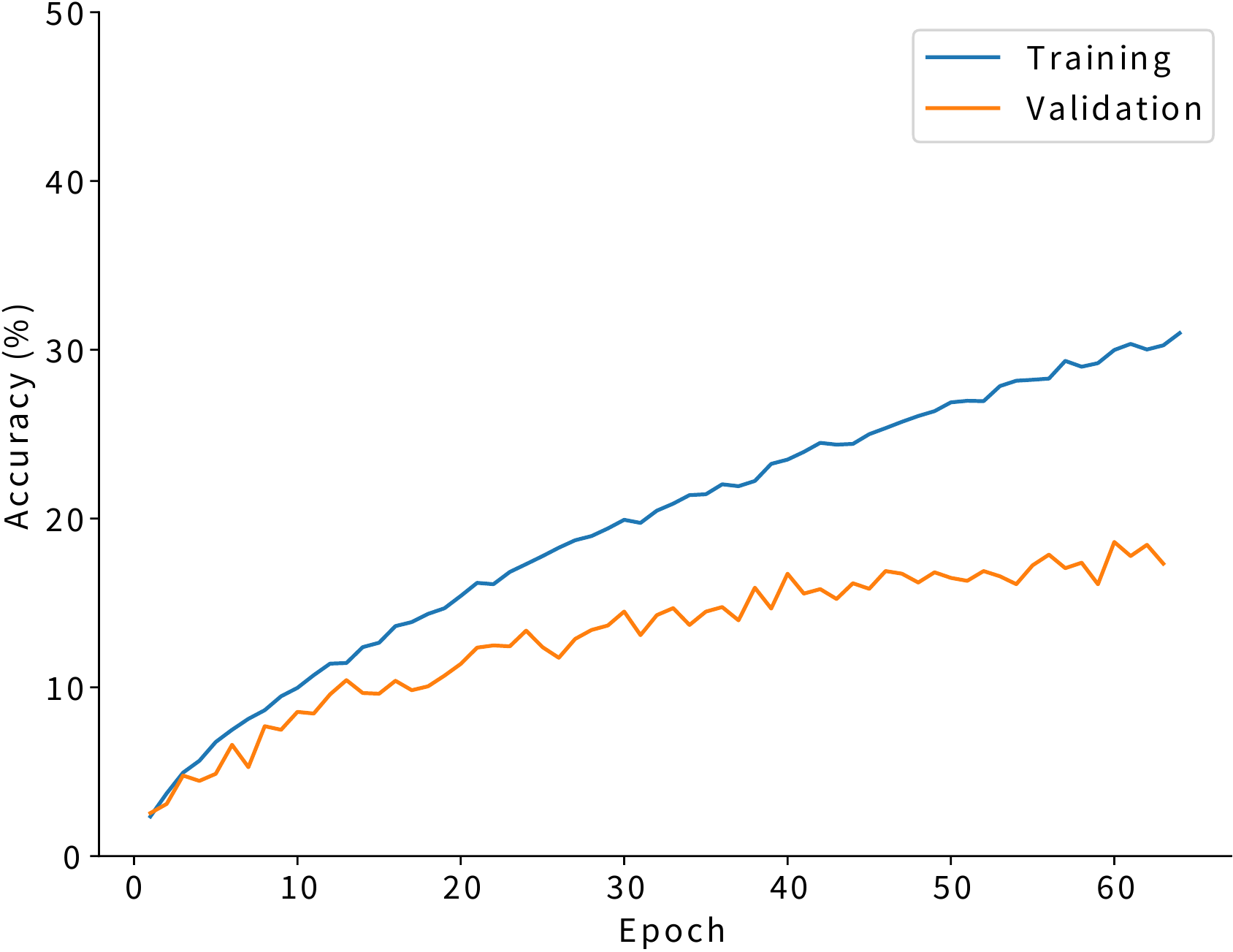}
      \caption{3D ResNet-18}
      \label{fig:training_activitynet_resnet}
    \end{subfigure}
    \begin{subfigure}[b]{0.5\linewidth}
      \includegraphics[width=\linewidth]{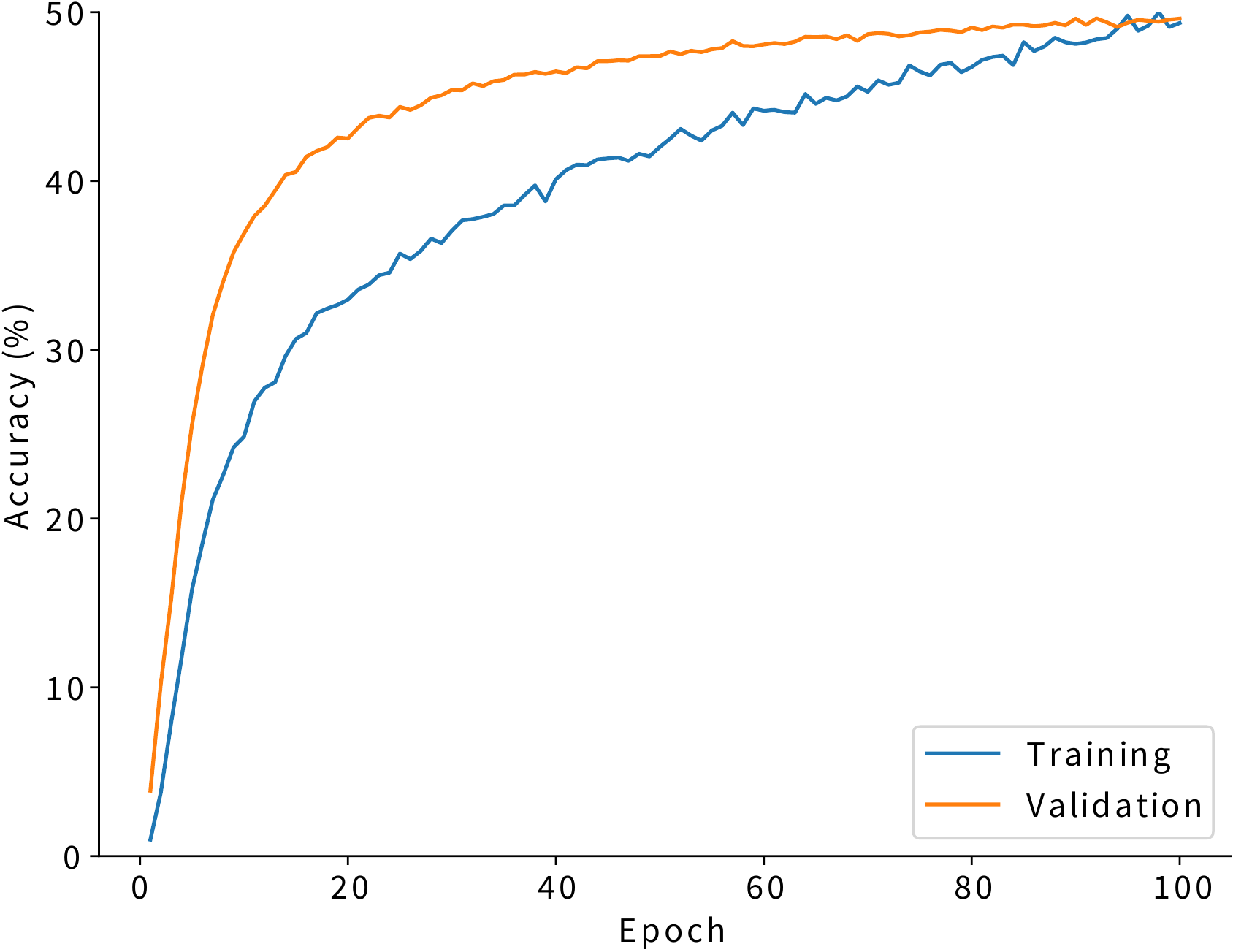}
      \caption{Sports-1M pretrained C3D}
      \label{fig:training_activitynet_c3d}
    \end{subfigure}
    \caption{
      Training of the models on the ActivityNet dataset.
      The size of ActivityNet is relatively small (20,000 videos) compared with the Kinetics (300,000 videos) and Sports-1M (1,000,000).
      The 3D ResNet overfitted because of the relatively small size whereas the C3D got better accuracies without overfitting.
    }
    \label{fig:training_activitynet}
  \end{figure*}
  \subsection{Dataset}
    In the experiments, we used the ActivityNet (v1.3) \cite{activitynet} and Kinetics datasets \cite{Kinetics}.
    The ActivityNet dataset provides samples from 200 human action classes with an average of 137 untrimmed videos per class and 1.41 activity instances per video.
    The total video length is 849 hours,
    and the total number of activity instances is 28,108.
    The dataset is randomly split into three different subsets:
    training, validation and testing,
    where 50\% is used for training, and 25\% for validation and testing.

    The Kinetics dataset has 400 human action classes,
    and consists of 400 or more videos for each class.
    The videos were temporally trimmed, so that they do not include non-action frames,
    and last around 10 seconds.
    The total number of the videos is 300,000 or over.
    The number of training, validation, and testing sets are about 240,000, 20,000, 40,000, respectively.

    The number of activity instances of the Kinetics is ten times larger than that of the ActivityNet
    whereas the total video lengths of the both datasets are close.

    For both datasets, we resized the videos to 360 pixels height without changing their aspect ratios, and stored them.

  \subsection{Results}
    We first describe the preliminary experiment on the ActivityNet dataset.
    The purpose of this experiment is exploring the training of the 3D ResNets on the relatively small dataset.
    In this experiment, we trained 18-layer 3D ResNet described in Table \ref{tbl:network} and Sports-1M pretrained C3D \cite{C3D}.
    Figure \ref{fig:training_activitynet} shows the training and validation accuracies in the training.
    The accuracies were calculated based on recognition of not entire videos but 16 frame clips.
    As shown in Figure \ref{fig:training_activitynet} \subref{fig:training_activitynet_resnet},
    the 3D ResNet-18 overfitted so that its validation accuracies was significantly lower than the training ones.
    This result indicates that the ActivityNet dataset is too small to train the 3D ResNets from scratch.
    By contrast, Figure \ref{fig:training_activitynet} \subref{fig:training_activitynet_c3d} shows that
    the Sports-1M pretrained C3D did not overfit and achieved better recognition accuracy.
    The relatively shallow architecture of the C3D and pretraining on the Sports-1M dataset
    prevent the C3D from overfitting.

\begin{figure*}[t]
      \begin{subfigure}[b]{0.5\linewidth}
        \includegraphics[width=\linewidth]{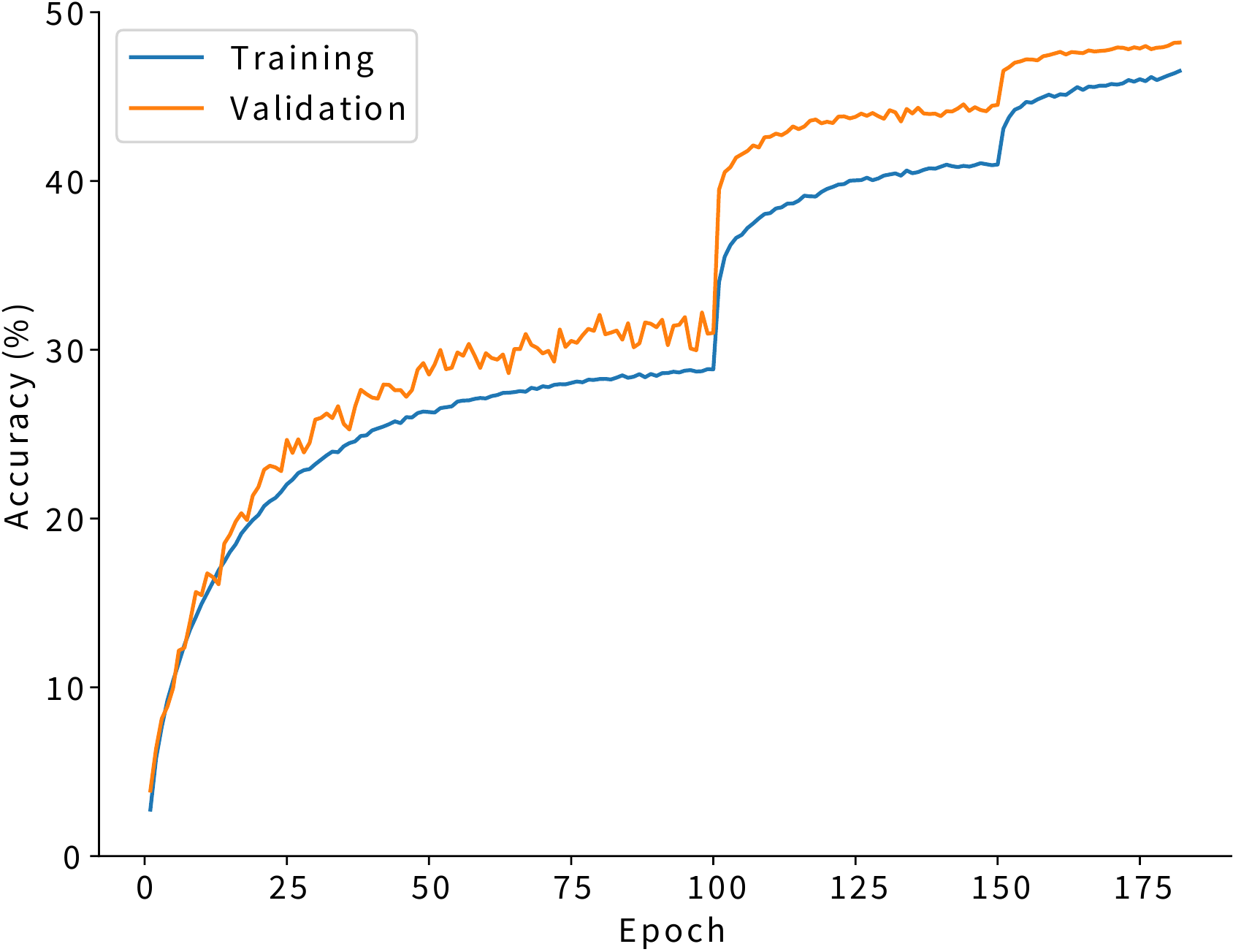}
        \caption{3D ResNet-34}
        \label{fig:training_kinetics_resnet}
      \end{subfigure}
      \begin{subfigure}[b]{0.5\linewidth}
        \includegraphics[width=\linewidth]{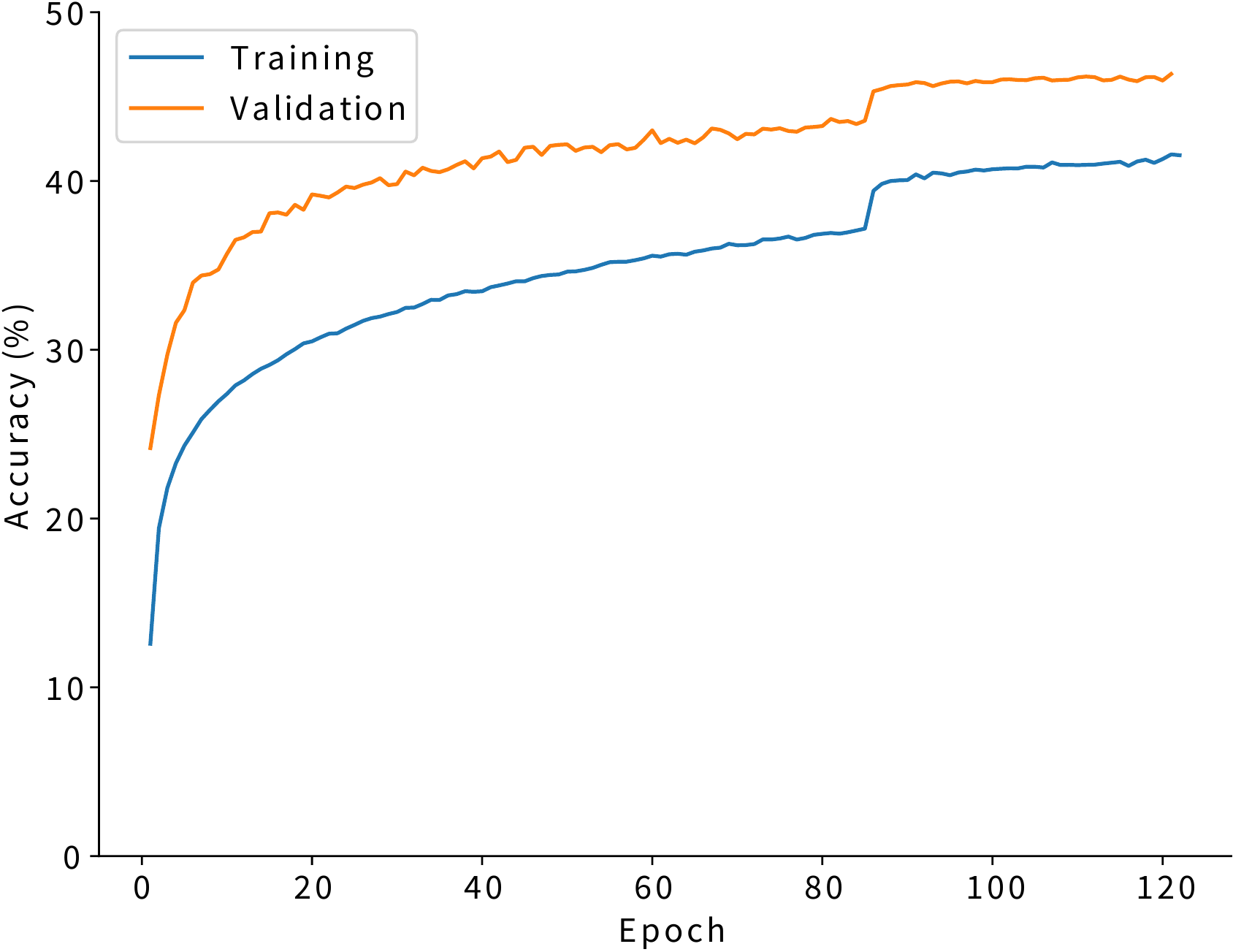}
        \caption{Sports-1M pretrained C3D}
        \label{fig:training_kinetics_c3d}
      \end{subfigure}
      \caption{
        Training of the models on the Kinetics dataset.
        The 3D ResNet achieved good performance without overfitting because of using the large-scale Kinetics dataset.
      }
      \label{fig:training_kinetics}
    \end{figure*}

    We then show the experiment on the Kinetics dataset.
    Here, we trained 34-layer 3D ResNet instead of 18-layer one
    because the number of activity instances of the Kinetics is significantly larger than that of the ActivityNet.
    Figure \ref{fig:training_kinetics} shows the training and validation accuracies in the training.
    The accuracies were calculated based on recognition of 16 frame clips, similar to Figure \ref{fig:training_activitynet}.
    As shown in Figure \ref{fig:training_kinetics} \subref{fig:training_kinetics_resnet},
    the 3D ResNet-34 did not overfit and achieved good performance.
    The Sports-1M pretrained C3D also achieved good validation accuracy, as shown in Figure \ref{fig:training_kinetics} \subref{fig:training_kinetics_c3d}.
    Its training accuracy, however, was clearly lower than the validation accuracy, (i.e. the C3D underfitted).
    In addition, the 3D ResNet is competitive to the C3D without pretraining on the Sports-1M dataset.
    These results indicate that the C3D is too shallow and the 3D ResNets are effective when using the Kinetics dataset.

    \begin{table*}[t]
        \centering
        \captionsetup{width=0.65\linewidth}
        \caption{
          Accuracy on the Kinetics dataset.
          \textit{Average} is averaged over Top-1 and Top-5 accuracies.
          * indicates the method performs pretraining on the Sports-1M dataset.
          Our 3D ResNet achieved higher accuracies than the C3D, which has relatively shallow architecture.
        }
        \label{tbl:results}
        \begin{tabular}{M{25mm}cccccc}
            \toprule
            \multirow{3}{*}{Method} & \multicolumn{6}{c}{Accuracy} \\
            \cmidrule{2-7}
            & \multicolumn{3}{c}{Validation set} & \multicolumn{3}{c}{Testing set} \\
            \cmidrule(l){2-4} \cmidrule(l){5-7}
            & Top-1 & Top-5 & Average & Top-1 & Top-5 & Average \\
            \midrule
            3D ResNet-34 (ours) & 58.0 & 81.3 & \textbf{69.7} & -- & -- & \textbf{68.9} \\
            C3D* & 55.6 & 79.1 & 67.4 & -- & -- & -- \\
            C3D w/ BN \cite{I3D} & -- & -- & -- & 56.1 & 79.5 & 67.8 \\
            RGB-I3D w/o ImageNet \cite{I3D} & -- & -- & -- & 68.4 & 88.0 & \textbf{78.2} \\
            \bottomrule
        \end{tabular}
    \end{table*}

    Table \ref{tbl:results} shows accuracies of our 3D ResNet-34 and state-of-the-arts.
    \textit{C3D w/ BN} \cite{Kinetics} is the C3D that employ batch normalization after each convolutional and fully connected layers.
    \textit{RGB-I3D w/o ImageNet} \cite{I3D} is the inception \cite{Inception}, which is very deep network (22 layers) similar to the ResNets,
    -based CNNs with 3D convolutional kernels.
    Here, we show the results of the RGB-I3D without pretraining on the ImageNet.
    The ResNet-34 achieved higher accuracies than Sports-1M pretrained C3D and C3D with batch normalization trained from scratch.
    This result supports the effectiveness of the 3D ResNets.
    By contrast, RGB-I3D achieved the best performance whereas
    the number of depth of ResNet-34 is greater than that of RGB-I3D.
    A reason for this result might be the difference of number of used GPUs.
    Large batch size is important to train good models with batch normalization \cite{I3D}.
    Carreira et al. used 32 GPUs to train the RGB-I3D whereas we used 4 GPUs with 256 batch size.
    They might use more large batch size on their training
    and it contribute to the best performance.
    Another reason might be the difference of sizes of input clips.
    The size for the 3D ResNet is $3 \times 16 \times 112 \times 112$ due to the GPU memory limits whereas that for the RGB-I3D is $3 \times 64 \times 224 \times 224$.
    High spatial resolutions and long temporal durations improve recognition accuracy \cite{LongTermTemporalConv}.
    Therefore, using a lot of GPUs and increasing batch size, spatial resolutions, and temporal durations might achieve further improvements of 3D ResNets.

\begin{figure*}[t]
      \centering
      \includegraphics[width=0.72\linewidth, clip]{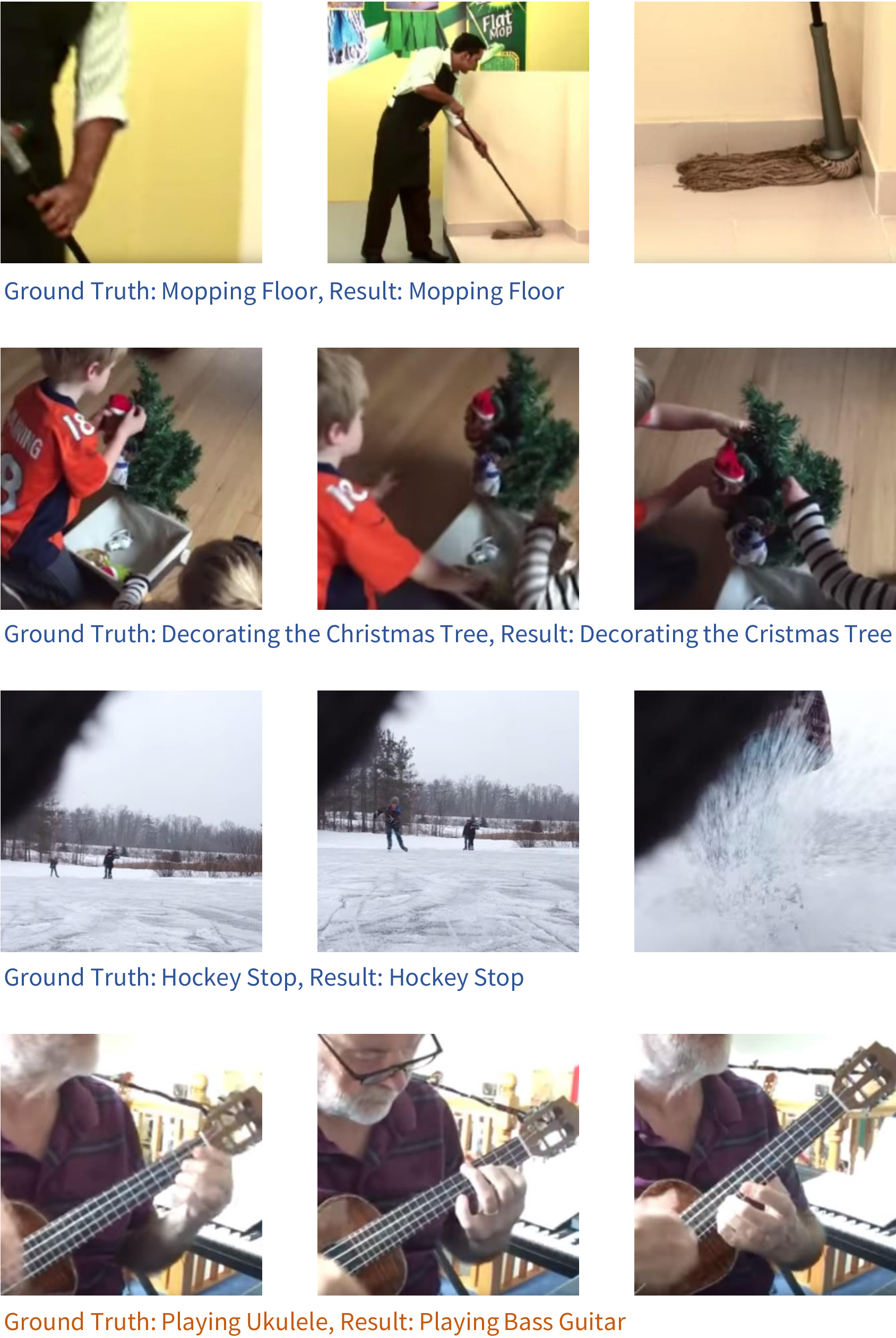}
      \captionsetup{width=0.72\linewidth}
      \caption{
        Examples of recognition results of 3D ResNets-34 on the Kinetics.
        The frames of each row are cropped at center positions
        and show part of the original videos.
        The three top rows are correctly recognized results.
        The bottom row is a wrongly recognized result.
      }
      \label{fig:examples}
    \end{figure*}

    Figure \ref{fig:examples} shows examples of classification results of 3D ResNets-34.

\section{Conclusion}
  We explore the effectiveness of ResNets with 3D convolutional kernels.
  We trained the 3D ResNets using the Kinetics dataset, which is a large-scale video datasets.
  The model trained on the Kinetics performs good performance without overfitting despite the large number of parameters of the model.
  Our code and pretrained models are publicly available at
  \href{https://github.com/kenshohara/3D-ResNets}{https://github.com/kenshohara/3D-ResNets}.

  Because of the very high computational time of the training of 3D ResNets (three weeks),
  we mainly focused on the ResNets-34.
  In future work, we will conduct additional experiments for deeper model (ResNets-50, -101)
  and other deep architectures, such as DenseNets-201 \cite{densenets}.

{\small
\bibliographystyle{ieee}

}

\end{document}